\documentclass[12pt]{spieman}  
\usepackage{amsmath,amsfonts,amssymb}
\usepackage{graphicx}\graphicspath{ {images/} }

\usepackage{setspace}
\usepackage{tocloft}

\usepackage{float}

\usepackage{microtype, booktabs, multicol, multirow}
\usepackage{subcaption}
\usepackage[scientific-notation=true]{siunitx}
\usepackage{hyperref}

\newcommand{\etal}{\textit{et al.}}

\title{Scout-Net: Prospective Personalized Estimation of CT Organ Doses from Scout Views}

\author[b,*]{Abdullah-Al-Zubaer Imran}
\author[a]{Sen Wang}
\author[c]{Debashish Pal}
\author[c]{Sandeep Dutta}
\author[d]{Bhavik Patel}
\author[a]{Evan Zucker}
\author[a]{Adam Wang}
\affil[a]{Stanford University, Department of Radiology, Stanford, CA 94305, USA}
\affil[b]{University of Kentucky, Department of Computer Science, Lexington, KY 40506, USA}
\affil[c]{GE HealthCare, Menlo Park, CA 94025, USA}
\affil[d]{Mayo Clinic, Scottsdale, AZ 85259, USA}

\cftpagenumbersoff{figure}
\cftpagenumbersoff{table}

\begin{document} 
\maketitle

\begin{abstract}

\noindent{\bf Purpose:} Estimation of patient-specific organ doses is required for more comprehensive dose metrics, such as effective dose. Currently, available methods are performed retrospectively using the CT images themselves, which can only be done after the scan. To optimize CT acquisitions before scanning, rapid prediction of patient-specific organ dose is needed prospectively, using available scout images. We, therefore, devise an end-to-end, fully-automated deep learning solution to perform real-time, patient-specific, organ-level dosimetric estimation of CT scans. 

\noindent{\bf Approach:} We propose the ``Scout-Net'' model for CT dose prediction at six different organs as well as for the overall patient body, leveraging the routinely obtained frontal and lateral scout images of patients, before their CT scans. To obtain reference values of the organ doses, we used Monte Carlo simulation and 3D segmentation methods on the corresponding CT images of the patients. 

\noindent{\bf Results:} We validate our proposed Scout-Net model against real patient CT data and demonstrate the effectiveness in estimating organ doses in real-time (only 27 ms on average per scan). Additionally, we demonstrate the efficiency (real-time execution), sufficiency (reasonable error rates), and robustness (consistent across varying patient sizes) of the Scout-Net model.

\noindent{\bf Conclusions:} An effective, efficient, and robust Scout-Net model, once incorporated into the CT acquisition plan, could potentially guide the automatic exposure control for balanced image quality and radiation dose.

\end{abstract}


{\noindent \footnotesize\textbf{*}Abdullah-Al-Zubaer Imran,  \linkable{aimran@uky.edu} }

\begin{spacing}{2}   

\section{Introduction}
\label{sec:intro}
Thanks to the technological improvements across hardware and software, computed tomography (CT) is one of the most frequently used imaging modality for clinical practice and facilitates countless diagnoses and image-based medical procedures \cite{lell2020recent, withers2021x}. The clinical utility has considerably increased especially with numerous advances---improvements in rotation time and detector size, helical scanning, advanced reconstruction, energy-integrating detector to photon-counting detector \cite{lell2020recent, willemink2018photon}. Nevertheless, CT also accounts for a vast amount of ionizing radiation exposure at the population level, especially with the rapid growth in CT examinations \cite{lell2020recent}. A general trade-off exists between dose and image quality as the CT images are acquired with the minmax objectives---maximize image quality while minimizing radiation dose \cite{kachelriess2020possible, imran2021ssiqa}. 

\begin{figure}
    \centering
    \includegraphics[width=0.6\linewidth, trim={2.5cm 2.25cm 2.5cm 2.25cm}, clip]{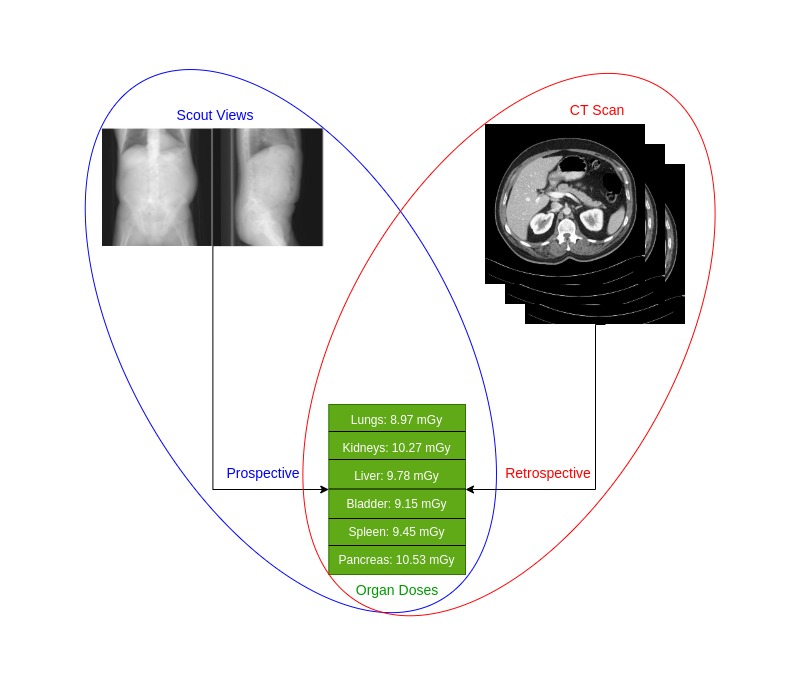}
    \caption{Personalized estimation of CT organ doses: \emph{retrospective} using the CT data itself; and \emph{prospective} from the scout views before performing the CT exams. Our goal is to predict the same organ dose distribution with the prospective approach as in the retrospective one.}
    \label{fig:overview}
 \end{figure}

Despite all the successes in CT imaging, the exposure of patients to ionizing radiation remains a major concern. Although radiation dose can be reduced to comply with the so called as low as reasonably achievable (ALARA) principle, it causes problems, such as image noise and artifacts \cite{imran2021ssiqa, imran2022personalized}. Therefore, it is extremely important to minimize dose while maintaining diagnostic value (maximized image quality). The baseline dose levels and the generated image quality could be established through the optimization of CT exam protocols \cite{damilakis2021ct}. But the current dose optimization mechanisms do not take patient-specific anatomy and the shape variation of organs into account. Patient-specific organ dose enables radiation risk metrics such as \emph{effective dose}, which weights the different organs by their relative radio-sensitivities based on the ICRP tissue weights \cite{valentin20072007}. Customization based on such metrics could enable patient-specific acquisitions that apply radiation exposure in the areas where it is most needed for the expected image quality and any potential clinical tasks, while avoiding organs more sensitive to the radiation, thereby reducing the stochastic risk.

There has been a large body of work on CT dosimetric quantification. The widely used CT dose index (CTDI) and dose length product (DLP) assess scanner output rather than patient dose. Size-specific dose estimate (SSDE) factors in patient size, but does not provide patient-specific organ dose. Prior works in organ dose have focused mostly on estimating CT doses either based on phantom data or from patient CT scans retrospectively (Figure~\ref{fig:overview}). Monte Carlo (MC) simulation is the gold-standard for accurately estimating patient dose maps. MC simulations can produce and track particles at the voxel level, and the patient-specific absorbed dose is calculated from the deposited energy per voxel \cite{furhang1996monte}. Through the years, these simulations have been accelerated through GPU computation, which parallelizes simulation of x-ray photon transport in voxel geometries \cite{badal2009accelerating, jia2012fast, xu2015archer}. Other approaches such as a GPU-based deterministic solver of the linear Boltzmann transport equation (LBTE) also offer accelerated performance \cite{wang2019fast, principi2020deterministic, principi2021validation}. While such approaches have made patient-specific dose estimation faster, they require 3D patient data and can only be performed retrospectively. Moreover, they also need to be paired with organ segmentation to obtain organ dose, with further limitation associated with obtaining high quality experts' annotation (segmentation references), computational cost, execution speed, etc. 

Data-driven automated CT dose estimation has recently gained a lot of attraction due to the advent of deep learning, particularly convolutional neural networks (CNNs). Several deep learning-based CT dose estimation approaches are available in the literature \cite{fan2020data, gotz2020deep, guerreiro2021deep, kontaxis2020deepdose, lee2019deep, maier2022real, zhu2020preliminary}. Maier \etal~proposed a dose estimation method based on 3D U-Net for reproducing the MC dose distribution from an input volumetric CT data \cite{maier2022real}. Peng \etal~obtained rapid patient-specific organ dose estimates by combining deep-learning-based multi-organ segmentation with GPU-accelerated MC dose maps \cite{peng2020method}, while Fu \etal~used individualized computational phantoms (``digital twins") for multi-organ segmentation \cite{fu2021iphantom}. Offe \etal~developed a fully-automated and rapid tool for patient-specific organ dose calculation from pediatric CT scans using a LBTE solver and a V-Net segmentation model \cite{offe2020evaluation}. Klein \etal~used patient-specific organ dose estimates to determine a tube current curve that minimizes a patient risk measure of dose such as effective dose while keeping the image quality constant \cite{klein2022patient}. However, all of these studies are based on the common assumption of the availability of a 3D patient representation. This can only be possible \emph{after} the CT scan is already obtained. If we aim to optimize the CT scan itself for an individual, these organ doses must be estimated \emph{before} acquiring the CT scan and also with the real patient data.

\begin{figure}
    \centering
    \includegraphics[width=\linewidth]{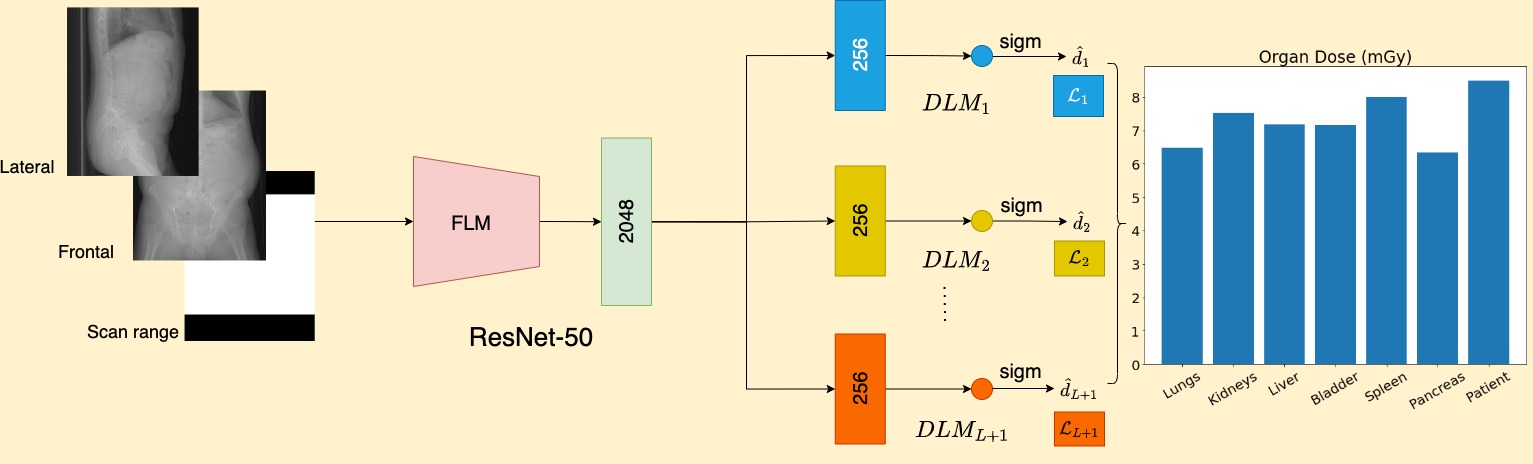}
    \caption{Schematic of the proposed Scout-Net model: The input scout views (frontal and lateral) are passed along with the scan range as 3-channel input to the generic feature learning module (FLM) based on the pretrained ResNet-50 model. The extracted features from FLM are passed to each of the $L$ organ-specific dose learning modules (DLMs). Organ dose prediction errors $\mathcal{L}_i$ ($i\in L$) are calculated separately at the DLMs. $\mathcal{L}_{L+1}$ denotes the loss for the patient body dose prediction via DLM$_{L+1}$. The bar chart shows the predicted doses obtained after de-normalization to the original scale.}
    \label{fig:model}
\end{figure}

Several studies in the literature have focused on estimating CT organ doses prospectively. Tian \etal~developed an atlas-based organ dose estimation method (matching the patient to a computational phantom) \cite{tian2015prospective}. In another patient-matching approach, Gao \etal~performed patient-specific organ and effective dose estimates for a large cohort of patients \cite{gao2020patient}. However, these methods are based on phantom data and rely on the accurate matching of the patient to a limited library of phantoms. Estimation of real patient data is desirable for realistic dose prediction.

There have been several proposed approaches for creating a 3D representation from scout views \cite{montoya2022reconstruction, shen2019patient, shapira2022convolutional}. While such scout--CT generation enables predicting organ doses in a prospective manner, in addition to the initial generation, the predicted CT-based processing (segmentation, dose estimation, etc.) incurs computational overhead. Moreover, the performance of the dose prediction could be degraded depending on the quality of the predicted 3D representations \cite{ying2019x2ct, almeida2021three}.

Instead, we propose predicting CT organ doses directly from scout images, in a prospective manner without requiring any additional processes related to 3D CT data. CT scout images are preliminary radiographs that are acquired primarily to assist in planning CT scans, such as for identifying the scan range, patient positioning, and tube current modulation. The scouts contribute only about 4\% of the radiation dose of a typical CT scan and can also contribute diagnostic information \cite{brook2007ct}. They have also been used for automatic image registration of a computational phantom for estimating CT organ dose \cite{kortesniemi2012organ}. While organ-based tube current modulation is available from some vendors, it is limited to certain organs, requires manual selection of organ locations, and follows prescribed rules for reducing tube current at certain angles (e.g., anterior direction for eye lens or female breasts), without accounting for patient-specific organ doses \cite{duan2011dose, wang2012bismuth, gandhi2015phantom}.

\subsection{Contributions}
Deviating from the existing CT dose estimation methods, we propose a simple yet effective, fully-automated, end-to-end CNN-based solution for patient-specific, organ dose estimation in real time from only the scout views. To our best knowledge, this is the first such study focused on estimating organ doses directly from scouts, even before acquiring the actual CT scans. Our Scout-Net model is capable of predicting patient-specific organ doses prospectively in real-time from scout images (lateral and frontal) only. To compute the reference CT organ doses, we configure a Monte Carlo dose engine (MC-GPU) for realistic CT system incorporating bowtie filtration and anode heel effect, in order to calculate CT dose maps of patients for a non-isotropic X-ray source. To obtain the organ doses, we apply the segmented organ masks from our 3D auto-segmentation tool to the dose maps. Our 3D auto-segmentation is based on a Context Encoder U-Net with weighted focal cross-entropy loss and can automatically segment the organs-of-interest from chest-abdomen-pelvis CT scans \cite{imran2021personalized}.

\subsection{Our Previous Work}
This paper is an extended version of our conference paper ``Personalized CT Organ Dose Estimation from Scout Images'' \cite{imran2021personalized}. In this manuscript, we substantially extend by adding a thorough literature review, modified Scout-Net architecture leveraging transfer learning with a pre-trained ResNet architecture, experiments with larger data size, additional experiments and baselines, improved results, more detailed description of the methods and results/discussion, and additional figures and visualizations.

\subsection{Paper Organization}
This paper is organized as follows: Sec.~\ref{sec:intro} discusses the background and motivation of the prospective CT organ dose estimation from scouts and deviation from existing methods in the literature; Sec.~\ref{sec:scout-net} introduces the proposed Scout-Net model for organ dose estimation from input scout views; Sec.~\ref{sec:dose_reference} illustrates the reference organ dose calculation from CT scans leveraging MC-GPU and automatic 3D segmentation (Detailed description of the segmentation model and results can be found in our conference paper \cite{imran2021personalized}); Sec.~\ref{sec:experiments} describes the materials used for the experiments and details of the models' implementations. Sec.~\ref{sec:results} demonstrates the dose prediction performance by the proposed Scout-Net and discusses the key findings at different experimental settings. And finally, Sec.~\ref{sec:conclusion} concludes by summarizing the contributions of the paper and future work for further improvements.

\section{Patient-Specific Organ Dose from Scouts}
\label{sec:scout-net}

Figure~\ref{fig:overview} gives an overview of CT organ dose estimation, either from scout views or from CT images. Either could be used, but existing methods use the CT images to retrospectively estimate organ dose. In order to perform the prospective dose estimates, our goal is to do it only using the scout views, avoiding reliance on CT images. 

\subsection{The Scout-Net Model}

With the goal of prospectively estimating personalized CT organ dose, we propose the Scout-Net model illustrated in Figure~\ref{fig:model}. The Scout-Net model is capable of predicting (mean) organ doses from input scout images (lateral and frontal views). Important information about patient size and anatomy are available from the scouts. Scouts are usually longer than the actual scans. A scan range input is therefore used along with the scouts to better inform the model on the scanned region in the scouts. With this additional information, the model becomes aware of where the patient will be scanned, which directly exposes those regions to radiation, while other regions are indirectly exposed through scattered radiation. Along with the patient-specific organ dose prediction, the model also facilitates predicting patient-specific body dose. 

The Scout-Net model comprises of two basic modules: a generic feature learning module (FLM) and multiple dose learning modules (DLMs). The FLM is used to extract shared features, later utilized through separate organ-specific DLMs. The model also estimates the patient's overall body dose through a separate DLM. In this paper, we refer to our previous scout-based dose prediction \cite{imran2021personalized} as Scout-Net-O. Scout-Net takes lateral scout ($x_{ls}$), frontal scout ($x_{fs}$), and scan range ($x_{sr}$) as inputs; and predicts organ/body doses $\hat{d}_l$ where $l\in {1,2,\dots,L+1}$. The inputs $x_{ls}, x_{fs}, x_{sr}$ are all concatenated to make a 3-channel input to the model. 

The customized layers in the Scout-Net-O model are replaced by the ResNet-50 \cite{he2016deep} architecture originally designed and trained on the large ImageNet dataset. ResNet-50 is a 50-layer deep CNN, effectively used as the backbone for many computer vision and medical imaging tasks. Since the FLM module in Scout-Net is used for feature extraction, the final fully-connected (1000-class classification) layer and softmax are removed from the ResNet-50 architecture. First, 3-channel scout inputs are fed to a convolutional layer with 64 filters with a kernel size of $7\times7$, which is followed by a max pooling layer of stride 2. The output of the max pooling layer is then passed through 5 layers, each comprising of $1\times1$, $3\times3$, and $1\times1$ convolutions and skip connections, with 3,4,6, and 3 repetitions respectively. With an average pool, the representation is obtained at 2048 feature dimension. 

The extracted features are then shared across the DLMs. Each DLM consists of two fully-connected layers (256 and 1 neurons respectively), a leakyReLU activation with slope 0.2, and finally a sigmoid to output the mean dose prediction in the normalized (0--1) scale. The Scout-Net models consists of $L+1$ dose learning modules (DLMs), where each individual DLM is a separate network with the objective of predicting the dose at an organ-of-interest. In addition to the $L$ organs, the ($L+1$)th DLM predicts the patient body dose. Based on the reference doses $d_l$ and the model predicted doses $\hat{d}_l$ at organ-of-interest $l$, the loss for the Scout-Net model at $DLM_l$ is therefore calculated as:
\begin{equation}
\label{eqn:dose_loss}
\begin{aligned}
\mathcal{L}^{dose}_{(d, \hat{d})} = \frac{1}{M}\sum_i^M \sum_l^{L+1} \| d_l{(i)} - \hat{d}_l{(i)} \|_2,
\end{aligned}
\end{equation}
where $\mathcal{M}$ denotes the minibatch size and the organ label $l$ is extended to cover the patient body label ($l\in\{1,2,\dots,L+1\}$) at $L+1$.

\begin{figure}
    \centering
    \includegraphics[width=0.7\linewidth]{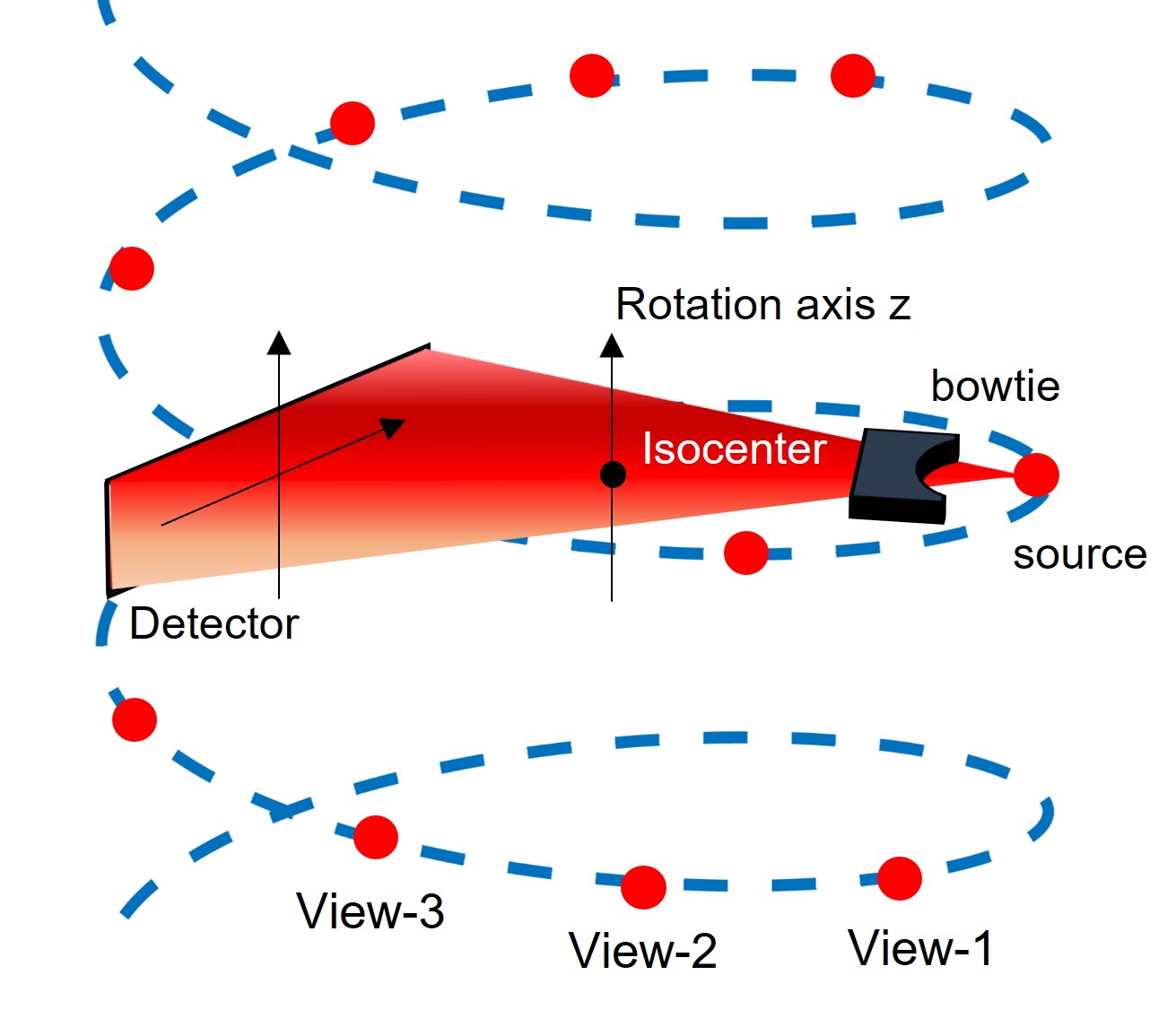}
    \caption{CT system with helical acquisition. Source positions are sampled discretely along the helical trajectory.}
    \label{fig:ct_system}
\vspace{0.75cm}   
\end{figure}

\begin{figure}
    \centering
    \resizebox{\linewidth}{!}{
    \subcaptionbox{Bowtie model}{\includegraphics[width=0.49\linewidth, trim={7cm 12cm 7cm 12cm},clip]{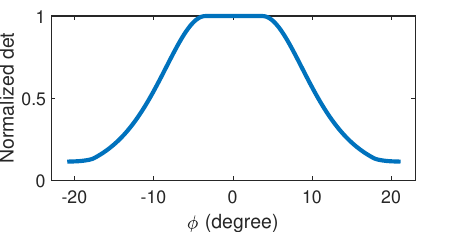}}
    \subcaptionbox{Anode heel effect}{\includegraphics[width=0.49\linewidth, trim={7cm 12cm 7cm 12cm},clip]{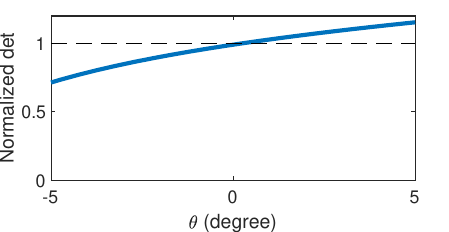}}
    }
    \caption{MC-GPU configuration: Normalized intensity (a) after bowtie filtration along the fan angle; and (b) along the cone angle, demonstrating the anode heel effect.}
    \label{fig:bt-heel}
\vspace{0.75cm}   
\end{figure}

\section{Organ Dose Estimation With CT Scans}
\label{sec:dose_reference}

In order to establish the reference organ doses, we still need the (real patient) CT data. Therefore, we calculate the patient-specific CT organ doses from CT scans. We utilized the open-source MC-GPU\footnote{\url{https://github.com/DIDSR/MCGPU}, version 1.3, accessed on August 10, 2020.} \cite{badal2009accelerating} and computed the 3D dose map by modeling more realistic CT scanners. Leveraging organ segmentation masks through a 3D CNN-based model \cite{imran2021personalized}, organ-specific dose scores are calculated. Previous studies have verified the robustness of CT organ doses against small errors in organ segmentation \cite{offe2020evaluation}, hence justifying the use of an automatic segmentation tool to avoid laborious and time-consuming expert-provided annotations.

\subsection{MC-GPU: Monte Carlo Dose Estimation}
\label{sec:mcdose}
MC simulations based on 3D patient representations (voxelized models) can provide personalized estimates of CT dose. The MC-GPU tool, validated independently on an AAPM reference phantom \cite{sharma2019real}, gives massive acceleration to the dose estimation. 


\begin{table}[t]
    \centering
    \caption{List of the customized parameters used to configure the MC-GPU dose calculation engine for our  realistic CT system.$^*$}
    \medskip
    \label{tab:mcgpu_config}
    \begin{tabular}{@{} lc c @{}}
    \toprule
    Parameter & \phantom{a} &  Choice(s)\\
    \midrule
    Tube potential  && 120 kVp\\
    Z-axis coverage && 80 mm\\
    Pitch factor && 0.99\\
    Voxel spacings (isotropic) && (4, 4, 4) mm$^3$\\
    Source to rotation axis distance && 625.61 mm\\
    Source to detector distance && 1097.61 mm\\
    Vertical translation between projections && 3.3 mm\\
    Polar and azimuthal apertures && ($42.0^\circ$, $7.32^\circ$) \\
    Angle between projections && $15^\circ$\\
    Views per rotation && 24\\
    Tube current (fixed) && 100 mAs\\
    \bottomrule
    \multicolumn{2}{c}{$^*$\footnotesize{Excluding the default MC-GPU parameters}}
    \end{tabular}
\vspace{0.75cm}
\end{table}


\begin{figure}[t]
    \centering
    \resizebox{0.8\linewidth}{!}{
    \begin{tabular}{l l l}
        & {\Huge Density map (g/cc)} & {\Huge Organ dose map (mGy)}
        \smallskip\\
        \rotatebox{90}{\hspace{2cm} \Huge Axial} & 
        \includegraphics[width=0.5\linewidth, trim={0cm 0cm 9.8cm 0cm},clip]{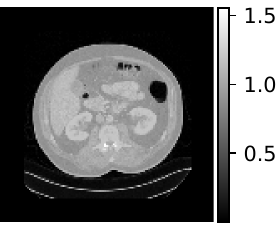}
        &
        \includegraphics[width=0.5\linewidth, trim={9.85cm 0cm 0cm 0cm},clip]{organ_dose_map_axial_1.pdf}
        \bigskip \\
        \rotatebox{90}{\hspace{1cm} \Huge Sagittal} &
        \includegraphics[width=0.6\linewidth, trim={0cm 0cm 9.7cm 0cm}, clip]{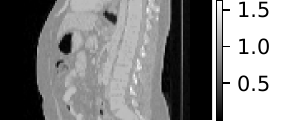}
        &
        \includegraphics[width=0.6\linewidth, trim={9.7cm 0cm 0cm 0cm}, clip]{organ_dose_map_sagittal_1.pdf}
        \bigskip \\
        \rotatebox{90}{\hspace{1cm} \Huge Coronal} &
        \includegraphics[width=0.6\linewidth, trim={0cm 0cm 9.7cm 0cm}, clip]{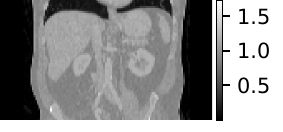}
        &
        \includegraphics[width=0.6\linewidth, trim={9.7cm 0cm 0cm 0cm}, clip]{organ_dose_map_coronal_1.pdf}
        \end{tabular}
}
    \caption{Generation of organ-specific dose maps from a CT scan leveraging the MC-GPU dose engine and the multi-organ segmentation: visualization of the input CT density map ($\rho$) and the resultant corresponding organ dose maps ($O$) at different views.}
    \label{fig:organ_dose_map}
\vspace{0.75cm}   
\end{figure}

Figure~\ref{fig:ct_system} illustrates the overall CT system and helical scan trajectory. Towards modeling a realistic CT system, we modified the original version of MC-GPU by including the bowtie filter and anode heel effect. A bowtie filter is usually employed in CT scans to reduce unnecessary radiation dose to the periphery of a patient and to equalize the transmitted signal at the detector \cite{liu2013dynamic, mail2009influence}. Our bowtie model analytically attenuates the source spectrum (e.g., 120 kVp) based on the known bowtie composition as a function of fan angle (polar angle $\phi$ in MC-GPU). We calculated the filtered spectra and relative photon intensity distribution for all fan angles from the input source spectrum (Figure~\ref{fig:bt-heel}a). Scattered photons from the bowtie were ignored, as their contribution to patient dose is small. Furthermore, the anode heel effect leads to x-ray intensity variation in the cone angle (azimuthal angle $\theta$) that can be modeled as a probability function \cite{kusk2021anode} (Figure~\ref{fig:bt-heel}b). Therefore, to randomly select a photon direction and energy for MC simulation, we sample the cone and fan angles according to their distributions, then sample the photon energy from the fan angle-specific spectrum. 

For individualized patient organ dose, patient CT images were used to generate patient-specific voxelized phantoms that contain spatial maps of both material type and mass density. The CT images were first resampled to isotropic voxels of $4\times4\times4$ mm$^3$ to keep the computational demands reasonable. The density mapping was performed following a piece-wise linear curve which defines the densities of the mixture of water and bone \cite{wang2018acuros}. Figure~\ref{fig:organ_dose_map} visualizes the MC-GPU inputs of mass density and corresponding material map in different views (axial, sagittal, coronal) of a patient CT scan.

We simulated a helical scan from the most superior to most inferior slice for the geometry of a GE Revolution CT scanner. Table~\ref{tab:mcgpu_config} lists the modified set of parameters for our MC-GPU configuration. We repeated the MC-GPU dose simulation with uniformly spaced apart $N=4$ start angles $(\delta(i)=\{0,90,180,270\}^\circ)$. Considering the actual start angles cannot be controlled prospectively, therefore, we averaged the $N$ dose maps to obtain the dose map \cite{sharma2019real}. 
\begin{equation}
    \label{eqn:dosemap}
    D_{avg} = \frac{1}{N}\sum_{i=1}^N MCGPU(V, \Psi) |_{\delta(i)},
\end{equation}
where, $V$ is the voxelized phantom and $\Psi$ denotes the set of all the parameters used to configure the MC-GPU simulation. MC-GPU dose is reported as eV/g/photon and was scaled to the more standard mGy for a 100 mAs scan with constant tube current, using a scanner-specific calibration to convert a simulated 32 cm CT dose index (CTDI) phantom to a physical CTDI measurement (1 eV/g/photon = 1.8143 mGy/100 mAs). Figure~\ref{fig:organ_dose_map} visualizes a representative dose map. Using a threshold $t_{air} = 0.1$ g/cc in the voxelized phantom, we masked out the air and obtained the patient body dose map. Therefore, from average dose map in (\ref{eqn:dosemap}), the final dose map is obtained as
\begin{equation}
    \label{eqn:body_dose}
    D = D_{avg} \odot (V>t_{air}).
\end{equation}


\subsection{Volumetric Multi-Organ Segmentation}
Our multi-organ segmentation model leverages a 3D Context Encoder U-Net \cite{gu2019cenet, cciccek20163d} network. The context encoder utilizes atrous convolution at different rates at the end of the encoder block which enables capturing longer range context related information compared to the standard U-Net \cite{chen2017rethinking}. The decoder employs residual multi-kernel pooling which performs max-pooling at multiple field-of-views. The network is trained separately for the $L$ different organs, primarily in the thorax, abdomen, and pelvis regions.

Hyperparameter optimization of the model followed the process and learning from a previous study\cite{dutta2019assessment} to find the most optimal learning rate, optimizer, depth of the convolution blocks, and number of kernels. A 3D patch-based training was used with an input patch size of 256$\times$256$\times$8 voxels. Different min-max normalization values were used to pre-process the input for different organs in order to train the model with the range of Hounsfield Unit values appropriate for those organs.

The encoder and decoder networks utilize 5 convolution blocks followed by downsampling and upsampling respectively. The final convolution is followed by a softmax activation. The models were trained with a focal categorical cross-entropy loss \cite{lin2017focal}.

\begin{equation}
\label{eqn:seg_loss}
\begin{aligned}
L^{seg}_{(y, \hat{y})} = -\sum_t\sum_i\alpha(1 - \hat{y}_{t,i})^{\gamma} y_{t,i} \log(\hat{y}_{t,i}),
\end{aligned}
\end{equation}
where, $i$ iterates over the number of channels and $t$ iterates over the number of voxels in the image patches. 
A signed-distance weighted map was calculated to emphasize the voxels near the boundary of the reference and was added to the loss function in (\ref{eqn:seg_loss}) \cite{ronneberger2015unet}.

\begin{equation}
    \begin{aligned}
    w_t = a \cdot \exp{(-b \cdot e_t)} +1,
    \end{aligned}
\end{equation}
where, $e_t$ is the Euclidean distance from the closest label; weighting factors $a=2.0$ and $b=0.5$ were chosen for the experiments.

Each CT exam in the test set was inferenced by the selected organ model from the training process using overlapped patches. The entire CT volume was subdivided into successive 256$\times$256$\times$8 voxel patches and each patch was inferenced by the model at a time. At the end, results from all patches were combined to have a complete organ segmentation mask for that CT volume. The successive patches were extracted with an overlap of 50 percent along all three axes to improve segmentation performance at the edges of each individual patch. The voxels that were inferenced multiple times as part of different patches were adjudicated based on a majority vote. There is a trade-off between segmentation accuracy and inference time in the selection of the overlap interval, and 50 percent was found to be optimal in our experiment.

\begin{figure}[t]
    \centering
    \includegraphics[width=0.75\linewidth]{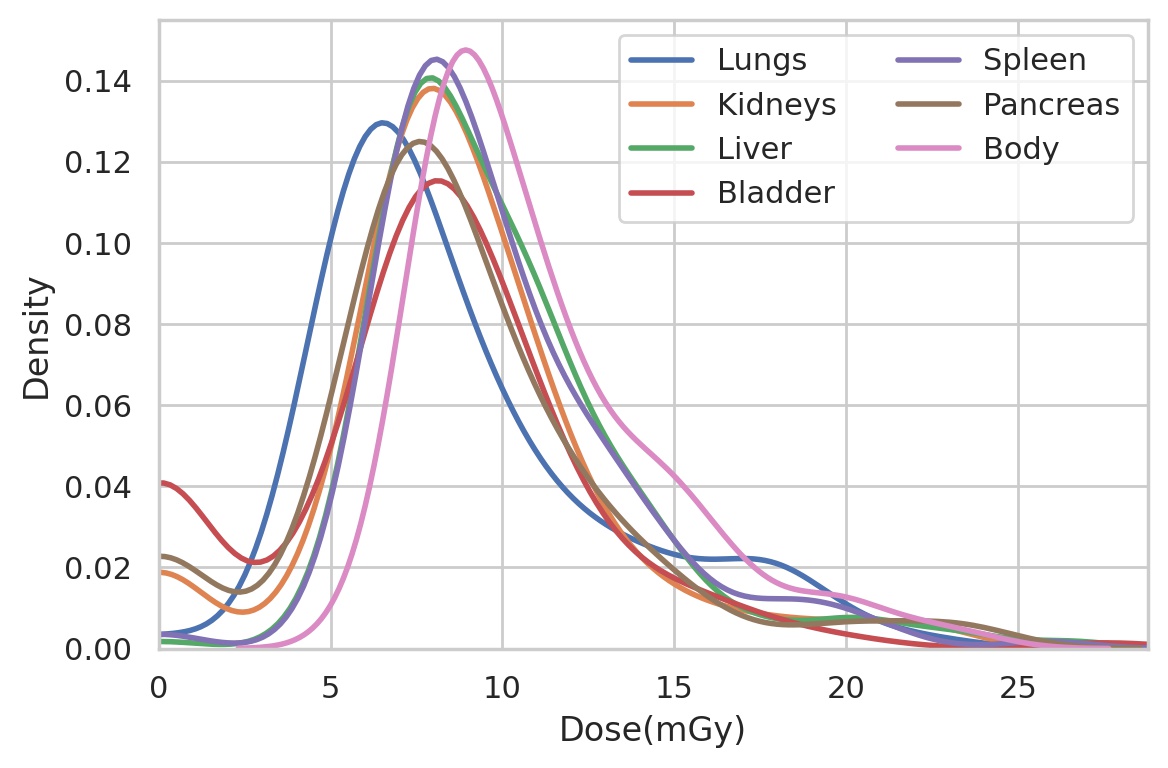}
    \caption{Distribution of organ and body doses of all patients in the dataset. Mean organ doses are calculated following (\ref{eqn:mean_dose}).}
    \label{fig:voxel_dose}
\vspace{0.75cm}   
\end{figure}

\subsection{Reference Organ Doses} 
In order to accurately calculate the CT doses at organs-of-interest, we applied the organ segmentation masks obtained from our 3D segmentation model. The segmentation output $(y)$ of a 3D CT scan was interpolated and resampled to align with the CT voxel coordinates following the geometry prepared for MC-GPU. The organ-specific dose map $O_l$, $l\in L$ was generated from the patient-specific dose map $D$ in (\ref{eqn:body_dose}).  
\begin{equation}
\label{eqn:organ_dose}
    O_l = D \odot y_l; l\in L.
\end{equation}


Following the ICRP publication 103 \cite{valentin20072007}, we then calculated the mean absorbed dose in a specific organ $l$ ($l\in L$) from the organ dose map in (\ref{eqn:organ_dose}). Since the organ-specific dose is assumed to be correlated with radiation detriment from stochastic effects across the whole organ, the mean organ dose ($d_l$) is therefore calculated by summing up the absorbed energy in the organ $l$ and then dividing by the mass of that organ.
\begin{equation}
    \label{eqn:mean_dose}
    d_{l} = \frac{\int_l D(a,b,c) \rho(a,b,c) dV}{\int_l \rho(a,b,c) dV}, 
\end{equation}
where $D(a,b,c)$ is the absorbed dose at a point $(a,b,c)$ in the organ $l$ and $\rho$ is the mass density at this point. Such mean absorbed doses in all the $L$ organs establish the basis of the dose quantification definition for protection \cite{protection2007icrp}.

\begin{figure}[t]
    \centering
    \includegraphics[width=0.9\linewidth]{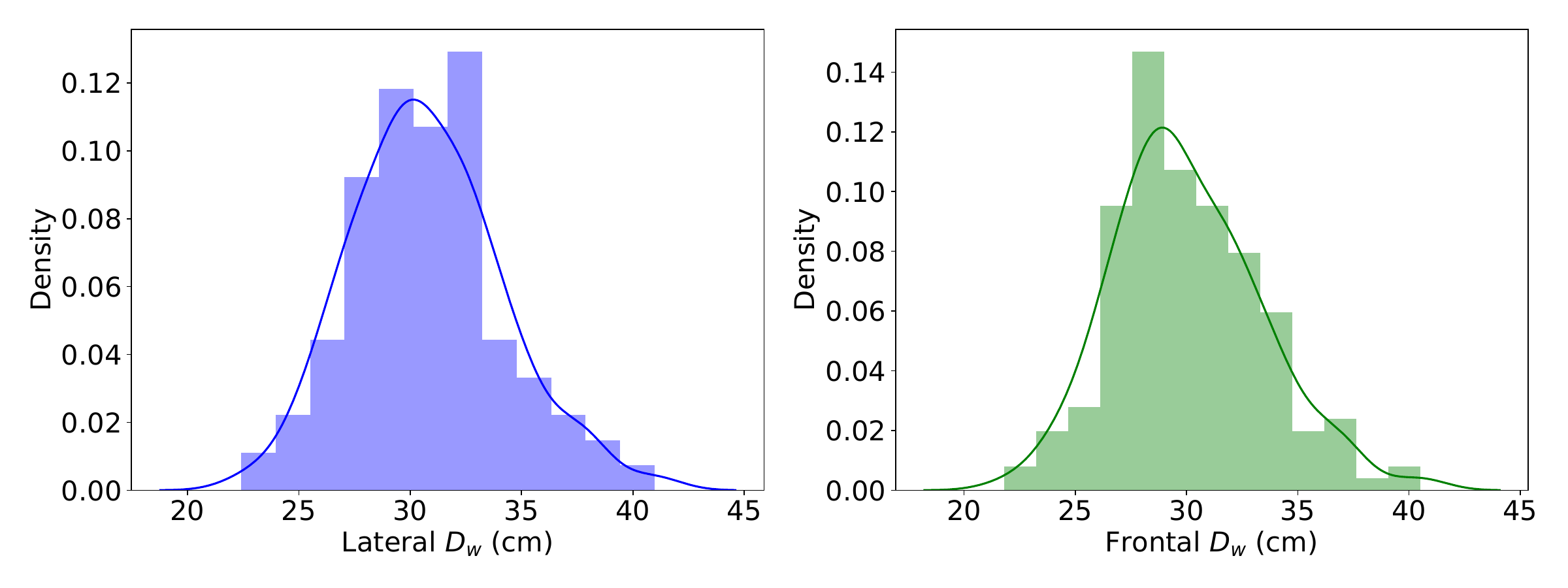}
    \caption{Distributions of patient (mean)  water-equivalent diameters ($D_w$) in both lateral and frontal scout views across the dataset: The histogram (discrete) bins show the counts in arbitrary ranges and the solid curves show the continuous estimation of the same distributions.}
    \label{fig:DWDist}
\vspace{0.75cm}   
\end{figure}

An example patient CT density map and the corresponding organ dosemap (segmented organ contours overlaid on the dosemap) at three different views obtained from the modified MC-GPU tool can be seen in Figure~\ref{fig:organ_dose_map}. Moreover, Figure~\ref{fig:voxel_dose} shows the distributions of organ and patient body doses for all the patient CT scans.

\section{Experimental Evaluations}
\label{sec:experiments}

\subsection{Materials}
\paragraph{Organ Dose Prediction:}We created a dataset by collecting 175 primarily contrast-enhanced body scans of adult patients.  The scans were acquired from Revolution CT scanners (GE Healthcare), and the dataset includes paired 2D (frontal and lateral) scout views and 3D CT images. Please note that the scout images do not contain any contrast. Due to patient motion in between the scouts and CT scan, there exist small inconsistencies, which is realistic also in the case of clinical use. All the scan data were collected with IRB (Institutional Review Board) approval from the participating institutions. The dataset has representative coverage of heterogeneous patient populations---adult patients, with patient water equivalent diameter ($D_w$) ranging from 21 cm to 41 cm (see Figure \ref{fig:DWDist}) \cite{mccollough2014use}.

\paragraph{Multi-Organ Segmentation:} We leveraged a combination of internal and publicly available CT datasets  
\footnote{\url{https://wiki.cancerimagingarchive.net/display/Public/Pancreas-CT}}
$^,$
\footnote{\url{https://www.synapse.org/\#!Synapse:syn3193805/wiki/89480}}
$^,$
\footnote{\url{http://medicaldecathlon.com/}} 
for the volumetric segmentation tasks. 
For consistency, we set the training size to 160 CT exams for every organ, that were obtained from up to five sites across different geographical regions, was used to train the organ segmentation models. The datasets were chosen to be varied in terms of the age, gender, size of the patients, presence and absence of iodinated contrast, presence and absence of pathology, as well as different reconstruction techniques and kernels. The organ boundaries were manually annotated by a team of trained radiologists. The expert annotations were used as the ground truth for training and for evaluating the model performance. The training dataset was further split into a training, validation, and evaluation dataset (70/10/20) for selecting the best performing model for each organ. In addition, another 45 CT exams that were not part of the training dataset were used as the independent test set to report the segmentation performance.

Leveraging the MC-GPU dose engine and the auto-segmentation tool \cite{imran2021personalized}, we calculated the doses at six different organs-of-interest: lungs (left and right), kidneys (left and right), liver, bladder, spleen, and pancreas.

\subsection{Implementation Details}

{\bf Input:} For the dataset, the scan range was first prepared at $690\times 530$ pixels ($1\times1$ mm$^2$ per pixel) and then the frontal and lateral scouts were registered to it. Therefore, inputs at $690\times 530\times3$ resolution are fed to the Scout-Net model instead of the original ResNet-50 input resolution $224\times 224\times3$). 

\noindent{\bf Hyper-parameters:} For training the model, we used the Adam optimizer and the learning rate was set to \num{2e-4}. The models were trained for 150 epochs, with a minibatch size of 16. 

\noindent{\bf Augmentations:} The training data were augmented by performing random vertical flip and Gaussian noise addition, to improve generalization and enlarge the scope of data distributions. Other basic augmentations could not be used as they are not physically consistent between the scout views and the CT scan. 

\noindent{\bf Baselines:} Unfortunately, there is no prior method available that could readily be used as a direct baseline. However, replacing the 2D operations in the Scout-Net-O model by their 3D counterparts, a retrospective CT-Net model was previously trained and compared \cite{imran2021personalized}. Here, we train the proposed Scout-Net model with the single scout views (Lateral or Frontal only) with the scan range as the 2-channel input while keeping all other settings the same and compared the results against our primary Scout-Net model. 

\noindent{\bf Computational Resource:} All the models were implemented in Python with the PyTorch framework and run on a \emph{Intel Core i7 64GiB} machine with an 8GB \emph{Nvidia GeForce RTX 2080 Super} GPU.

\noindent{\bf MC-GPU:} The MC-GPU simulation was modeled to simulate the GE Revolution CT scanner. For the sake of reference organ dose calculation, the modified MC-GPU code was compiled in C++/CUDA and wrapped in Python, and was run on the same machine as the Scout-Net model. 

\noindent{\bf Segmentation:} The segmentation model was trained for up to 50 epochs, and the best model was saved based on
validation accuracy. A Learning Rate Scheduler was used
for training the model. We used a minibatch size
of 16 and the Adam optimizer with an initial learning rate of
$0.0005$. The best model weights were saved for each of the
organ models and later used for separate inferences. The segmentation models were implemented in Python with the TensorFlow 2.1 framework, and were run on an \emph{Nvidia DGX} machine with an 16GB \emph{Nvidia V100} GPU.

\subsection{Evaluation Metric}
We employed the Dice similarity coefficient (Dice) for evaluating our 3D multi-organ segmentation model performance. For every organ $l \in L$, the Dice score is calculated from the target mask $y_l$ and model predicted mask $\hat{y}_l$:
\begin{equation}
    Dice = \frac{2\cdot|y_l \cap \hat{y}_l|} {|y_l| + |\hat{y}_l|}.
\end{equation}

We evaluated the organ dose prediction models by performing 5-fold cross validation and use two error measures to report the (organ/body-specific) dose predictions: mean relative percentage error (PE) and root-mean-square error (RMSE). From the reference dose scores $d_l$ and the predicted doses $\hat{d}_l$ at organ $l$, the PE error (\%) is calculated as:
\begin{equation}
    PE = \frac{|d_l - \hat{d}_l|}{d_l} \times 100\%.
\end{equation}

And the RMSE error (mGy) is calculated as:
\begin{equation}
\begin{aligned}
RMSE = \sqrt{\| d_l - \hat{d}_l \|^2_2}.
\end{aligned}
\end{equation}

\begin{table}[t]
    \centering
    \caption{Performance evaluation (Dice) of the proposed segmentation model in segmenting organs from chest-abdomen-pelvis CT scans.}
    \medskip
    \label{tab:seg_performance}
    \resizebox{0.7\linewidth}{!}{
    \begin{tabular}{cc cc cc cc cc cc c}
    \toprule
    \multirow{2}{*}{Q Score} & \phantom{a} & \multicolumn{11}{c}{Organs-of-interest}\\
    \cmidrule{3-13}
    && Liver && Lungs && Kidneys && Spleen && Pancreas && Bladder\\
    \midrule
    Q1 && 0.770 && 0.926 && 0.889 && 0.949 && 0.728 && 0.664 \\
    Q2 && 0.942 && 0.945 && 0.925 && 0.960 && 0.807 && 0.842 \\
    Q3 && 0.955 && 0.967 && 0.942 && 0.968 && 0.853 && 0.923 \\
    \bottomrule\\
    \end{tabular}
    }
\end{table}


\section{Results \& Discussion}
\label{sec:results}
We observe and analyze the performance of our Scout-Net model extensively on four different grounds: \emph{effectiveness} to verify how well the model can predict the organ doses, \emph{robustness} to investigate how well the model is generalized across different patients, \emph{sufficiency} to support the scout-based prospective dose prediction, and \emph{efficiency} to determine if the execution speed is fast enough (possibly real-time).

\begin{table}[t]
    \centering
    \setlength{\tabcolsep}{4pt}
    \caption{Relative percentage error (\%) comparison of the prospective Scout-Net against single view models in estimating the organ doses. Mean$\pm$stdev error rates are reported after cross-validation of the models.}
    \medskip
    \label{tab:perror}
    \resizebox{\linewidth}{!}{
    \begin{tabular}{l cc cccccc}
    \toprule
    \multirow{2}{*}{Model} & \multirow{2}{*}{Body} & \phantom{a} & \multicolumn{6}{c}{Organs-of-interest}\\
    \cmidrule{4-9}
    &&& Lungs & Kidneys & Liver & Bladder & Spleen & Pancreas\\
    \midrule
    Scout-Net &
    {\bf4.624$\pm$2.548} && 
    {\bf8.249$\pm$6.113} & {\bf6.023$\pm$3.053} & {\bf5.758$\pm$3.595} & {\bf8.980$\pm$6.852} & {\bf6.122$\pm$4.179} & {\bf6.480$\pm$3.859}\\
    Lateral &
     6.290$\pm$3.278 &&
    10.069$\pm$5.981 & 10.418$\pm$4.287 & 8.764$\pm$4.257 & 12.324$\pm$5.482 & 7.516$\pm$3.809 & 10.025$\pm$6.011\\ 
    Frontal &
    6.463$\pm$3.198 &&
    8.458$\pm$5.064 & 8.954$\pm$5.054 & 7.632$\pm$3.799 & 
    11.464$\pm$6.939 & 6.998$\pm$4.183 & 10.952$\pm$6.678\\ 
    \bottomrule
    \end{tabular}
    }
    \vspace{0.75cm}
\end{table}


\begin{figure}[t]
    \centering
    \includegraphics[width=\linewidth]{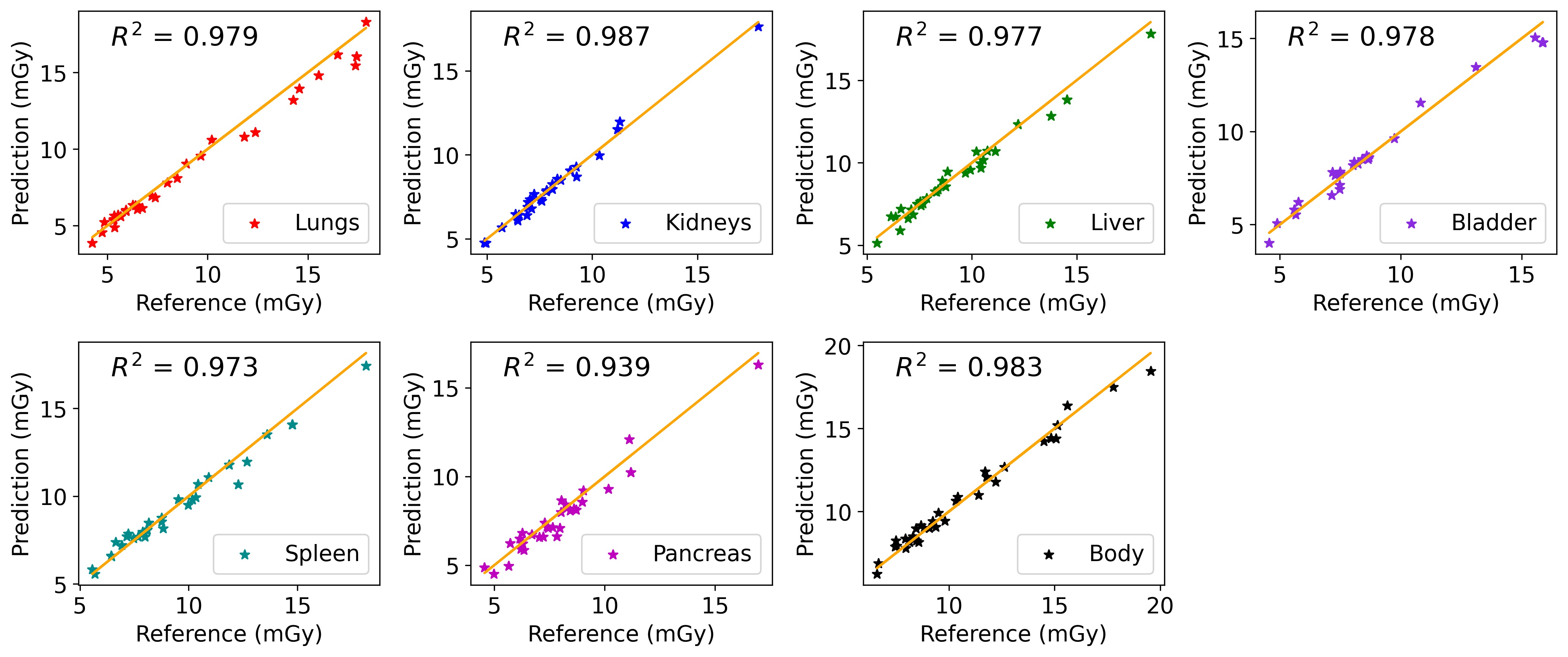}
    \caption{Scout-Net predicted organ doses compared to the reference doses. The identity line is plotted and its $R^2$ value is shown for each individual organ as well as patient's body.}
    \label{fig:gt_pred}
\vspace{0.75cm}   
\end{figure}

\subsection{Effectiveness}

Table~\ref{tab:seg_performance} details the segmentation performance in terms of the Dice scores. We report the quartile values for each of the individual organs-of-interest. Our 3D CT organ segmentation model achieved an average Dice score of 0.907 across the six organs, which is either superior or on par with several other existing multi-organ segmentation techniques \cite{wang2021region, lin2021variance}.

Automatic organ segmentation performance does not substantially impact the CT organ dose calculation as verified in a study by Offe \etal~\cite{offe2020evaluation}. Therefore, the model-generated 3D organ segmentations could be used for our reference dose calculations. Figure~\ref{fig:gt_pred} showcases the Scout-Net predicted organ doses for each of the six organs and patient body are in good agreement with the reference organ/body doses. The $R^2$ scores are consistently above 0.93 for all the organs.

\begin{figure}[t]
\resizebox{0.95\linewidth}{!}{
\subcaptionbox{Relative error comparison}{
\includegraphics[width=0.495\linewidth]{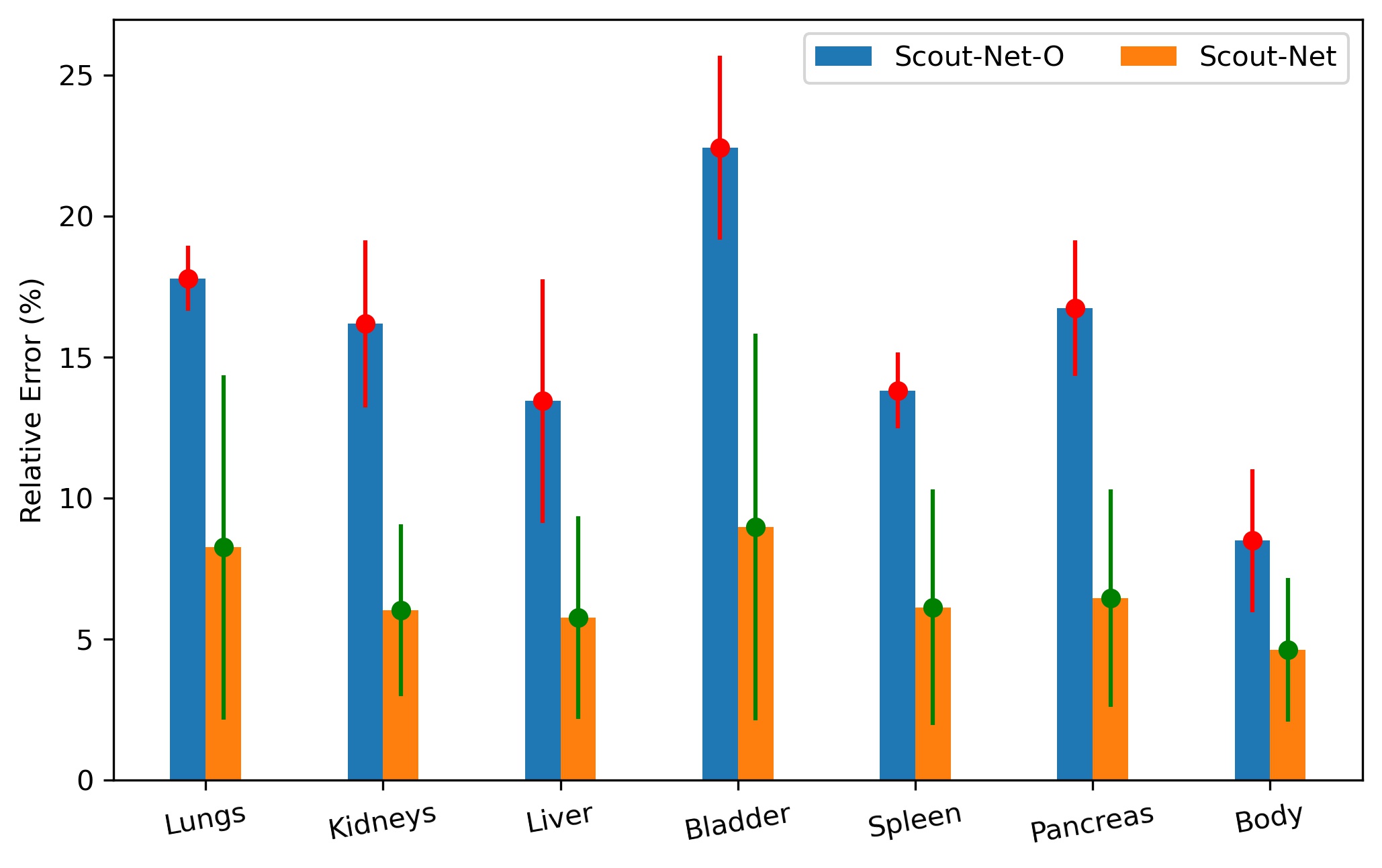}}
\subcaptionbox{RMSE comparison}{
\includegraphics[width=0.495\linewidth]{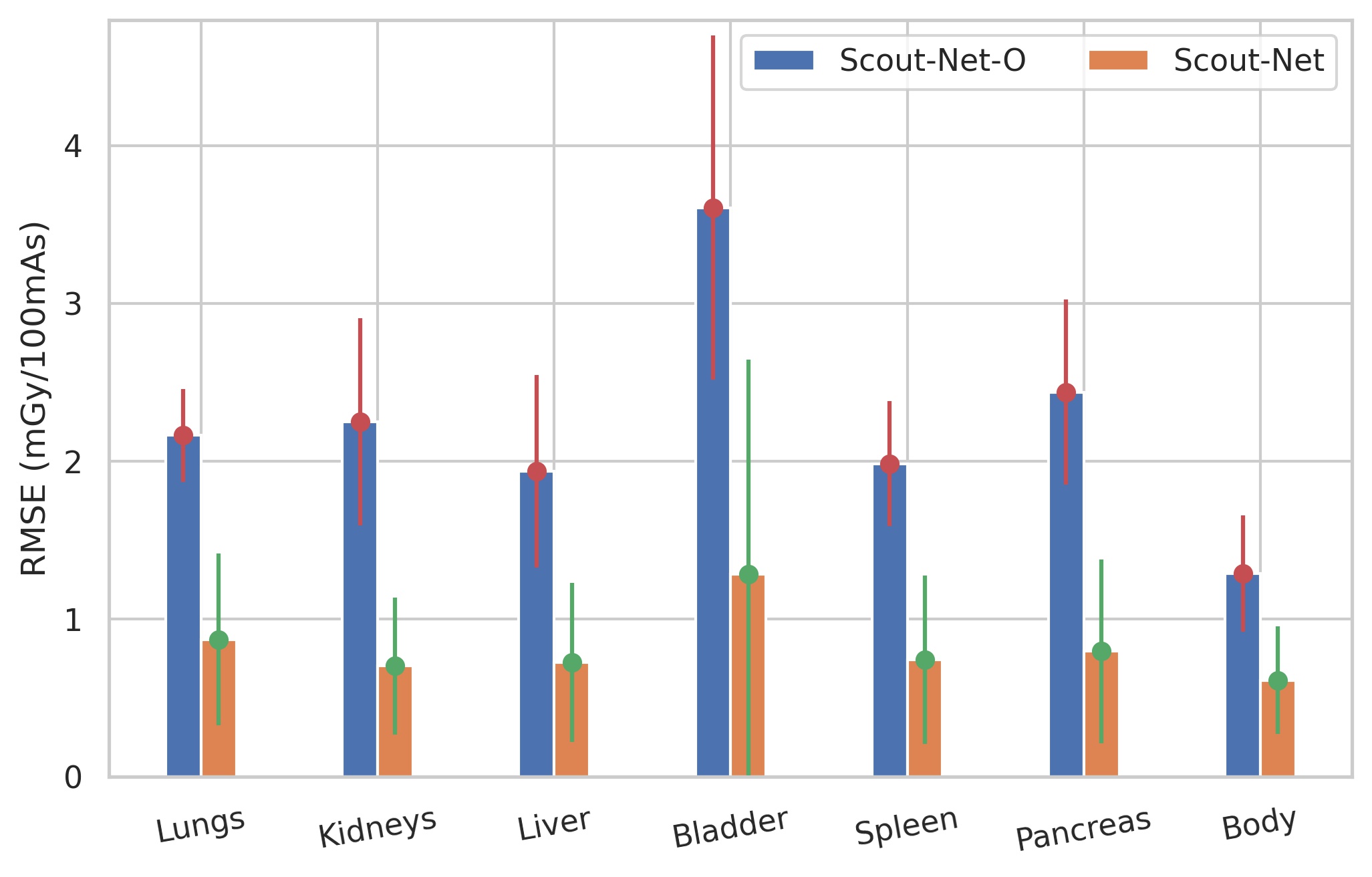}}
}
\caption{Relative error (\%) and RMSE comparisons on 5-fold cross-validation demonstrate superior  performance by our Scout-Net with transfer learning compared to the previous model (Scout-Net-O \cite{imran2021personalized}) in predicting the organ doses.}
\label{fig:vsscout*}
\vspace{0.75cm}  
\end{figure}

Tables~\ref{tab:perror} and \ref{tab:rmse_score} report the (percentage) relative error and RMSE error respectively for organ-/body-specific dose estimation. As can be seen, the proposed Scout-Net model with both lateral and frontal views obtained superior error rates in terms of both PE and RMSE over the baseline single-view models, achieving an average error rate of approximately 6.60\% and an average RMSE score of 0.819 mGy. Except in bladder (on par with the RMSE of Lateral only model), our Scout-Net achieves consistently better PE and RMSE scores for each of the organs as well as patient body. The model obtained error rates ranging between 4\% to 9\% for the organs and less than 5\% for the patient body. The superiority of the Scout-Net model over the single view (Frontal and Lateral only) models demonstrates the benefit of using both the views in predicting organ doses. Moreover, the Scout-Net model with transfer learning based on the pretrained ResNet-50 model, on average reduced errors in patient dose by approximately 45\% and about 55\% in organ doses from the previous Scout-Net-O model \cite{imran2021personalized}. The consistent improvement in dose prediction errors (PE) is observed in Figure \ref{fig:vsscout*}(a). Similar performance gain is observed in terms of the RMSE error measures across the organs and the patient body (Figure \ref{fig:vsscout*}(b)), demonstrating the effectiveness of using a pretrained ResNet-50 model as the FLM module. 


\begin{table}[t]
    \centering
    \setlength{\tabcolsep}{2pt}
    \caption{RMSE error (mGy) comparison of the prospective Scout-Net against single view models in estimating the organ doses. Mean$\pm$stdev error rates are reported after cross-validation of the models.}
    \medskip
    \label{tab:rmse_score}
    \resizebox{\linewidth}{!}{
    \begin{tabular}{l cc cccccc}
    \toprule
    \multirow{2}{*}{Model} & \multirow{2}{*}{Body} & \phantom{a} & \multicolumn{6}{c}{Organs-of-interest}\\
    \cmidrule{4-9}
    &&& Lungs & Kidneys & Liver & Bladder & Spleen & Pancreas\\
    \midrule
    Scout-Net & 
    {\bf0.613$\pm$0.340} && 
    {\bf0.871$\pm$0.543} & {\bf0.703$\pm$0.433} & {\bf0.725$\pm$0.504} & 1.284$\pm$1.359 & {\bf0.742$\pm$0.535} & {\bf0.796$\pm$0.583}\\
    Lateral &
    1.018$\pm$0.699 && 
    1.049$\pm$0.644 & 1.248$\pm$0.560 & 
    1.133$\pm$0.628 & {\bf1.270$\pm$0.562} & 
    0.949$\pm$0.545 & 1.257$\pm$0.847\\
    Frontal &
    0.918$\pm$0.398 && 
    0.895$\pm$0.532 & 1.039$\pm$0.696 & 
    0.948$\pm$0.545 & 1.386$\pm$0.997 & 
    0.843$\pm$0.522 & 1.319$\pm$1.015 \\
    \bottomrule
    \end{tabular}
    }
\vspace{0.75cm}
\end{table}

\begin{figure}[t]
\resizebox{\linewidth}{!}{
\subcaptionbox{$D_w$ vs body/organ doses}{
\includegraphics[width=0.495\linewidth]{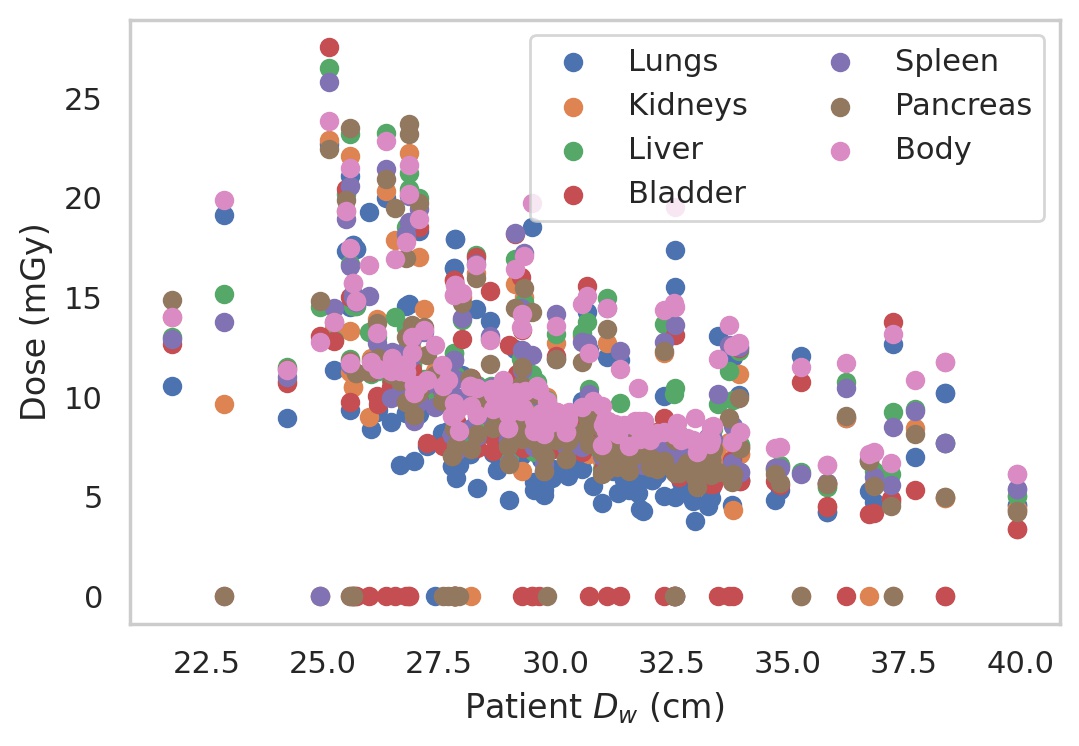}}
\subcaptionbox{$D_w$ vs relative dose error}{
\includegraphics[width=0.495\linewidth]{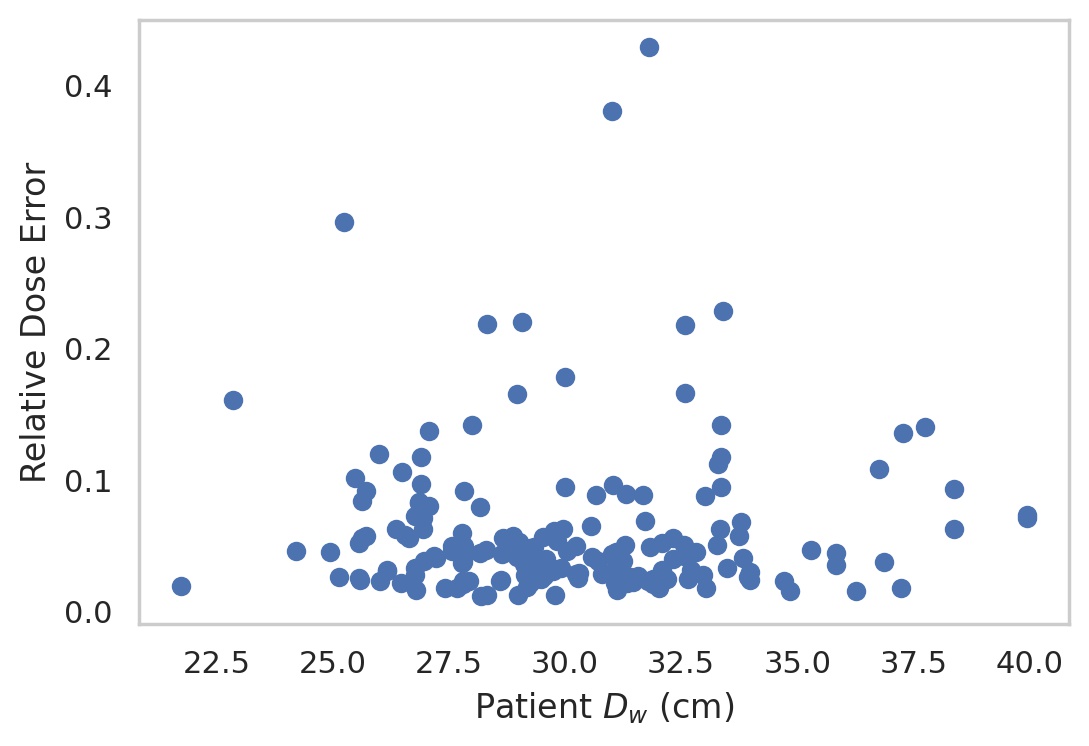}}
}
    \caption{Demonstration of robustness of our Scout-Net model for diverse patient sizes: (a) Mean $D_w$ (lateral and frontal) is strongly correlated with the dose across all the organs as well as patient body; (b) The model predicted doses are consistently at low error rates.}
    \label{fig:DWvsDose}
\vspace{0.75cm}   
\end{figure}

\begin{figure}
    \centering
    \includegraphics[width=0.6\linewidth]{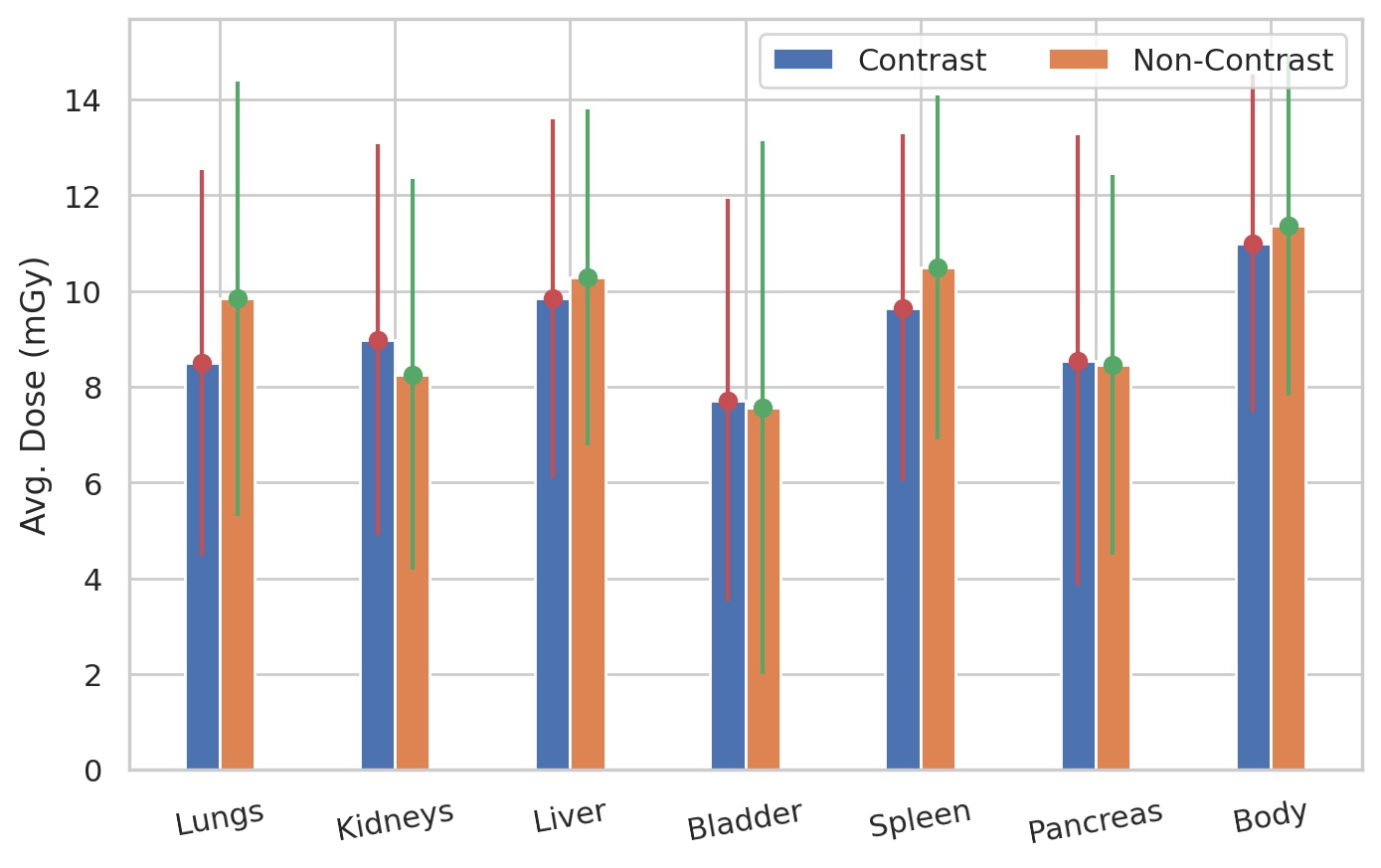}
    \caption{Side-by-side comparison of the dose scores reveals no significant differences between contrast and non-contrast CT scans: Average dose scores are shown for every organ as well as patient body.}
    \label{fig:ContrastDose}
\vspace{0.75cm}
\end{figure}

\subsection{Robustness}
For in-depth analysis of the Scout-Net results and the model robustness, we investigated organ doses based on patient $D_w$ and the use of contrast media. Patient $D_w$ is related to the size-specific dose estimates \cite{aapm2019size, juszczyk2021automated}, and we found strong correlations of the patient $D_w$ with doses of each of the organs as well as the patient body with $p< 0.001$ (see Figure~\ref{fig:DWvsDose}(a)). With the Spearman correlation, it was found that the organ dose prediction error is not correlated ($p> 0.05$) for any of the organs or patient body with the patient body size (see Figure~\ref{fig:DWvsDose}(b)). Considering the large $D_w$ coverage and heterogeneity in patient sizes in the dataset, this (no impact on dose prediction) demonstrates the robustness of our Scout-Net model in predicting the organ/body doses.

Moreover, we analyzed if the use of contrast media has any effect on the model performance. First, we categorized all our CT scans into contrast (129) and non-contrast (46) groups. We compared the reference doses across the individual organs and patient body. Figure~\ref{fig:ContrastDose} shows the side-by-side comparison of the contrast and non-contrast dose scores. No significant differences were observed as confirmed by performing the Wilcoxon signed rank test ($p> 0.05$). Similarly, we found there are no significant differences between relative errors in model predicted doses for contrast and non-contrast scans ($p> 0.05$). Therefore, it is confirmed that our proposed Scout-Net model is not biased to contrast enhancement. The similarities in dose could be because the contrast-enhanced scans were largely in the portal venous phase, and contrast was treated as dense water in our MC-GPU dose calculation, leading to similar dose values (increased energy deposition is normalized by increased mass density). Explicitly accounting for iodine as an additional material as well as other phases (e.g., arterial phase) may yield different doses that could be accounted for by Scout-Net as an input parameter to indicate whether the scan is with contrast (and its phase) or non-contrast.

\subsection{Sufficiency}
The Scout-Net-O model was previously found to be superior compared to the 3D CT-Net model in predicting the organ doses \cite{imran2021personalized}, likely due to our limited dataset size. The significant improvement ($p<0.05$) by the Scout-Net model over the Scout-Net-O further justifies the prospective estimation of CT organ doses from the scout views. For context, organ dose errors have been reported of up to 7\% due to imperfect organ segmentation \cite{schmidt2016accuracy}, up to 10\% when measured in physical phantoms with thermoluminescent dosimeters (TLDs) \cite{principi2021validation}, up to 15\% when comparing MC simulation against TLD measurements \cite{kalender2014generating}, and up to 20\% when using convolution-based estimation \cite{tian2016convolution}.

\subsection{Efficiency}
Another aim of any dose calculation algorithm is to maximize the calculation efficiency. In addition to robust and accurate estimation of the organ-specific and patient body doses, the proposed Scout-Net model also enables massive gain in the run-time with modest computing resources. The overall pipeline for the reference CT organ dose calculation from each CT scan takes about 200 seconds including preprocessing, the MC-GPU-based dose estimation, and segmentation of the six organs. On the other hand, with the proposed Scout-Net method, the CT organ dose estimation is about 7400$\times$ faster, taking only 27 ms for each scan. Moreover, Scout-Net has about a 560$\times$ speed gain over the retrospective CT-Net \cite{imran2021personalized} (approximately 15 s) in predicting patient-specific organ doses. In addition to the immense run-time advantage, our Scout-Net can even be utilized in limited computing settings as the CPU inference takes only 32 ms per scan which would ensure its wider applicability.

\section{Conclusions}
\label{sec:conclusion}
We have proposed a fully-automated, end-to-end, and real-time CT organ dose estimation method, namely Scout-Net. Our Scout-Net is capable of predicting patient-specific organ doses and patient body dose prospectively, even before doing the actual CT examinations. Additionally, to compute reference dose values, we have simulated more realistic CT dose estimation by modeling bowtie filtration and anode heel effect, via modification of the publicly available MC-GPU tool. We have developed an automated pipeline for calculating accurate reference CT organ doses by utilizing our 3D multi-organ segmentation tool with the MC-GPU dose calculation engine. Validating on real patient data with paired scouts and CTs, we have shown effective prediction of the prospective doses at six different organs-of-interest from the 2D scout views by our Scout-Net model. We have further demonstrated the efficiency (real-time execution), sufficiency (reasonable error rates), and robustness (consistent across varying patient sizes and contrast modes) of the Scout-Net model. Our ongoing effort includes extending the Scout-Net model to take tube current modulation \cite{klein2022patient, imran2022multimodal} and other scan parameters such as axial vs helical, helical pitch, tube voltage, bowtie filter, prefiltration, and gantry start angle into consideration, possibly as input parameters. Additionally, we are pursuing further validation of the model against a larger set of scans and improving the segmentation model for more accurate reference dose calculation towards improved dose prediction across the organs. We also envisage performing the CT dose estimate for additional organs-of-interest and at different CT scan locations. An effective, efficient, and robust Scout-Net model, once incorporated into the  CT acquisition plan, could potentially guide the automatic exposure control for balanced image quality and dose.  

\subsection* {Acknowledgments}
This work was supported by GE HealthCare.


\bibliography{report}   

\begin{thebibliography}{10}

\bibitem{lell2020recent}
M.~M. Lell and M.~Kachelrie{\ss}, ``Recent and upcoming technological developments in computed tomography: high speed, low dose, deep learning, multienergy,'' {\em Investigative radiology} {\bf 55}(1), 8--19  (2020).

\bibitem{withers2021x}
P.~J. Withers, C.~Bouman, S.~Carmignato, {\em et~al.}, ``X-ray computed tomography,'' {\em Nature Reviews Methods Primers} {\bf 1}(1), 1--21  (2021).

\bibitem{willemink2018photon}
M.~J. Willemink, M.~Persson, A.~Pourmorteza, {\em et~al.}, ``Photon-counting {CT:} technical principles and clinical prospects,'' {\em Radiology} {\bf 289}(2), 293--312  (2018).

\bibitem{kachelriess2020possible}
M.~Kachelrie{\ss} and M.~M. Rehani, ``Is it possible to kill the radiation risk issue in computed tomography?,'' {\em Physica Medica: European Journal of Medical Physics} {\bf 71}, 176--177  (2020).

\bibitem{imran2021ssiqa}
A.-A.-Z. Imran, D.~Pal, B.~Patel, {\em et~al.}, ``{SSIQA}: Multi-task learning for non-reference {{CT}} image quality assessment with self-supervised noise level prediction,'' in {\em 2021 IEEE 18th International Symposium on Biomedical Imaging (ISBI)},  1962--1965  (2021).

\bibitem{imran2022personalized}
A.-A.-Z. Imran, D.~Pal, S.~Wang, {\em et~al.}, ``Personalized ct organ noise estimation from scout images,'' in {\em Medical Imaging 2022: Physics of Medical Imaging},   {\bf 12031}, 186--192, SPIE  (2022).

\bibitem{damilakis2021ct}
J.~Damilakis, ``{CT Dosimetry:} what has been achieved and what remains to be done,'' {\em Investigative Radiology} {\bf 56}(1), 62--68  (2021).

\bibitem{valentin20072007}
J.~VALENTIN, ``The 2007 recommendations of the international commission on radiological protection. icrp publication 103,'' {\em Ann ICRP} {\bf 37}(2), 1--332  (2007).

\bibitem{furhang1996monte}
E.~E. Furhang, C.-S. Chui, and G.~Sgouros, ``A {Monte Carlo} approach to patient-specific dosimetry,'' {\em Medical physics} {\bf 23}(9), 1523--1529  (1996).

\bibitem{badal2009accelerating}
A.~Badal and A.~Badano, ``Accelerating {Monte Carlo} simulations of photon transport in a voxelized geometry using a massively parallel graphics processing unit,'' {\em Medical physics} {\bf 36}(11), 4878--4880  (2009).

\bibitem{jia2012fast}
X.~Jia, H.~Yan, X.~Gu, {\em et~al.}, ``Fast monte carlo simulation for patient-specific ct/cbct imaging dose calculation,'' {\em Physics in Medicine \& Biology} {\bf 57}(3), 577  (2012).

\bibitem{xu2015archer}
X.~G. Xu, T.~Liu, L.~Su, {\em et~al.}, ``Archer, a new monte carlo software tool for emerging heterogeneous computing environments,'' {\em Annals of Nuclear Energy} {\bf 82}, 2--9  (2015).

\bibitem{wang2019fast}
A.~Wang, A.~Maslowski, T.~Wareing, {\em et~al.}, ``A fast, linear boltzmann transport equation solver for computed tomography dose calculation (acuros ctd),'' {\em Medical physics} {\bf 46}(2), 925--933  (2019).

\bibitem{principi2020deterministic}
S.~Principi, A.~Wang, A.~Maslowski, {\em et~al.}, ``Deterministic linear boltzmann transport equation solver for patient-specific ct dose estimation: Comparison against a monte carlo benchmark for realistic scanner configurations and patient models,'' {\em Medical Physics} {\bf 47}(12), 6470--6483  (2020).

\bibitem{principi2021validation}
S.~Principi, Y.~Lu, Y.~Liu, {\em et~al.}, ``Validation of a deterministic linear boltzmann transport equation solver for rapid ct dose computation using physical dose measurements in pediatric phantoms,'' {\em Medical Physics}   (2021).

\bibitem{fan2020data}
J.~Fan, L.~Xing, P.~Dong, {\em et~al.}, ``Data-driven dose calculation algorithm based on deep {U-Net},'' {\em Physics in Medicine \& Biology} {\bf 65}(24), 245035  (2020).

\bibitem{gotz2020deep}
T.~I. G{\"o}tz, C.~Schmidkonz, S.~Chen, {\em et~al.}, ``A deep learning approach to radiation dose estimation,'' {\em Physics in Medicine \& Biology} {\bf 65}(3), 035007  (2020).

\bibitem{guerreiro2021deep}
F.~Guerreiro, E.~Seravalli, G.~Janssens, {\em et~al.}, ``Deep learning prediction of proton and photon dose distributions for paediatric abdominal tumours,'' {\em Radiotherapy and Oncology} {\bf 156}, 36--42  (2021).

\bibitem{kontaxis2020deepdose}
C.~Kontaxis, G.~Bol, J.~Lagendijk, {\em et~al.}, ``{DeepDose:} towards a fast dose calculation engine for radiation therapy using deep learning,'' {\em Physics in Medicine \& Biology} {\bf 65}(7), 075013  (2020).

\bibitem{lee2019deep}
M.~S. Lee, D.~Hwang, J.~H. Kim, {\em et~al.}, ``{Deep-dose:} a voxel dose estimation method using deep convolutional neural network for personalized internal dosimetry,'' {\em Scientific reports} {\bf 9}(1), 1--9  (2019).

\bibitem{maier2022real}
J.~Maier, L.~Klein, E.~Eulig, {\em et~al.}, ``Real-time estimation of patient-specific dose distributions for medical ct using the deep dose estimation,'' {\em Medical Physics} {\bf 49}(4), 2259--2269  (2022).

\bibitem{zhu2020preliminary}
J.~Zhu, X.~Liu, and L.~Chen, ``A preliminary study of a photon dose calculation algorithm using a convolutional neural network,'' {\em Physics in Medicine \& Biology} {\bf 65}(20), 20NT02  (2020).

\bibitem{peng2020method}
Z.~Peng, X.~Fang, P.~Yan, {\em et~al.}, ``A method of rapid quantification of patient-specific organ doses for {CT} using deep-learning-based multi-organ segmentation and {GPU-accelerated Monte Carlo} dose computing,'' {\em Medical physics} {\bf 47}(6), 2526--2536  (2020).

\bibitem{fu2021iphantom}
W.~Fu, S.~Sharma, E.~Abadi, {\em et~al.}, ``iphantom: a framework for automated creation of individualized computational phantoms and its application to ct organ dosimetry,'' {\em IEEE Journal of Biomedical and Health Informatics}   (2021).

\bibitem{offe2020evaluation}
M.~Offe, D.~Fraley, P.~M. Adamson, {\em et~al.}, ``Evaluation of deep learning segmentation for rapid, patient-specific {CT} organ dose estimation using an lbte solver,'' in {\em Medical Imaging 2020: Physics of Medical Imaging},   {\bf 11312}, 113124O, International Society for Optics and Photonics  (2020).

\bibitem{klein2022patient}
L.~Klein, C.~Liu, J.~Steidel, {\em et~al.}, ``Patient-specific radiation risk-based tube current modulation for diagnostic ct,'' {\em Medical Physics}   (2022).

\bibitem{tian2015prospective}
X.~Tian, X.~Li, W.~P. Segars, {\em et~al.}, ``Prospective estimation of organ dose in {CT} under tube current modulation,'' {\em Medical physics} {\bf 42}(4), 1575--1585  (2015).

\bibitem{gao2020patient}
Y.~Gao, U.~Mahmood, T.~Liu, {\em et~al.}, ``Patient-specific organ and effective dose estimates in adult oncologic {CT},'' {\em American Journal of Roentgenology} {\bf 214}(4), 738--746  (2020).

\bibitem{montoya2022reconstruction}
J.~C. Montoya, C.~Zhang, Y.~Li, {\em et~al.}, ``Reconstruction of three-dimensional tomographic patient models for radiation dose modulation in ct from two scout views using deep learning,'' {\em Medical Physics} {\bf 49}(2), 901--916  (2022).

\bibitem{shen2019patient}
L.~Shen, W.~Zhao, and L.~Xing, ``Patient-specific reconstruction of volumetric computed tomography images from a single projection view via deep learning,'' {\em Nature biomedical engineering} {\bf 3}(11), 880--888  (2019).

\bibitem{shapira2022convolutional}
N.~Shapira, S.~Bharthulwar, and P.~B. No{\"e}l, ``Convolutional encoder-decoder networks for volumetric computed tomography surviews from single-and dual-view topograms,'' {\em medRxiv}   (2022).

\bibitem{ying2019x2ct}
X.~Ying, H.~Guo, K.~Ma, {\em et~al.}, ``X2ct-gan: reconstructing ct from biplanar x-rays with generative adversarial networks,'' in {\em Proceedings of the IEEE/CVF conference on computer vision and pattern recognition},  10619--10628  (2019).

\bibitem{almeida2021three}
D.~F. Almeida, P.~Astudillo, and D.~Vandermeulen, ``Three-dimensional image volumes from two-dimensional digitally reconstructed radiographs: A deep learning approach in lower limb ct scans,'' {\em Medical Physics} {\bf 48}(5), 2448--2457  (2021).

\bibitem{brook2007ct}
O.~R. Brook, L.~Guralnik, and A.~Engel, ``{CT} scout view as an essential part of {CT} reading,'' {\em Australasian radiology} {\bf 51}(3), 211--217  (2007).

\bibitem{kortesniemi2012organ}
M.~Kortesniemi, E.~Salli, and R.~Seuri, ``Organ dose calculation in {CT} based on scout image data and automatic image registration,'' {\em Acta Radiologica} {\bf 53}(8), 908--913  (2012).

\bibitem{duan2011dose}
X.~Duan, J.~Wang, J.~A. Christner, {\em et~al.}, ``Dose reduction to anterior surfaces with organ-based tube-current modulation: evaluation of performance in a phantom study,'' {\em American journal of roentgenology} {\bf 197}(3), 689--695  (2011).

\bibitem{wang2012bismuth}
J.~Wang, X.~Duan, J.~A. Christner, {\em et~al.}, ``Bismuth shielding, organ-based tube current modulation, and global reduction of tube current for dose reduction to the eye at head ct,'' {\em Radiology} {\bf 262}(1), 191--198  (2012).

\bibitem{gandhi2015phantom}
D.~Gandhi, D.~J. Crotty, G.~M. Stevens, {\em et~al.}, ``Phantom study to evaluate the dose and image quality effects of a computed tomography organ-based tube current modulation technique,'' {\em Medical physics} {\bf 42}(11), 6572--6578  (2015).

\bibitem{imran2021personalized}
A.-A.-Z. Imran, S.~Wang, D.~Pal, {\em et~al.}, ``Personalized {CT} organ dose estimation from scout images,'' in {\em International Conference on Medical Image Computing and Computer-Assisted Intervention},  488--498, Springer  (2021).

\bibitem{he2016deep}
K.~He, X.~Zhang, S.~Ren, {\em et~al.}, ``Deep residual learning for image recognition,'' in {\em Proceedings of the IEEE conference on computer vision and pattern recognition},  770--778  (2016).

\bibitem{sharma2019real}
S.~Sharma, A.~Kapadia, W.~Fu, {\em et~al.}, ``A real-time {Monte Carlo} tool for individualized dose estimations in clinical {CT},'' {\em Physics in Medicine \& Biology} {\bf 64}(21), 215020  (2019).

\bibitem{liu2013dynamic}
F.~Liu, G.~Wang, W.~Cong, {\em et~al.}, ``Dynamic bowtie for fan-beam {CT},'' {\em Journal of X-ray Science and Technology} {\bf 21}(4), 579--590  (2013).

\bibitem{mail2009influence}
N.~Mail, D.~Moseley, J.~Siewerdsen, {\em et~al.}, ``The influence of bowtie filtration on cone-beam {CT} image quality,'' {\em Medical physics} {\bf 36}(1), 22--32  (2009).

\bibitem{kusk2021anode}
M.~W. Kusk, J.~M. Jensen, E.~H. Gram, {\em et~al.}, ``Anode heel effect: Does it impact image quality in digital radiography? a systematic literature review,'' {\em Radiography}   (2021).

\bibitem{wang2018acuros}
A.~Wang, A.~Maslowski, P.~Messmer, {\em et~al.}, ``Acuros {CTS:} {A} fast, linear {Boltzmann} transport equation solver for computed tomography scatter--part ii: System modeling, scatter correction, and optimization,'' {\em Medical physics} {\bf 45}(5), 1914--1925  (2018).

\bibitem{gu2019cenet}
Z.~{Gu}, J.~{Cheng}, H.~{Fu}, {\em et~al.}, ``Ce-net: Context encoder network for 2d medical image segmentation,'' {\em IEEE Transactions on Medical Imaging} {\bf 38}(10), 2281--2292  (2019).

\bibitem{cciccek20163d}
{\"O}.~{\c{C}}i{\c{c}}ek, A.~Abdulkadir, S.~S. Lienkamp, {\em et~al.}, ``{3D U-Net:} learning dense volumetric segmentation from sparse annotation,'' in {\em International conference on medical image computing and computer-assisted intervention},  424--432, Springer  (2016).

\bibitem{chen2017rethinking}
L.-C. Chen, G.~Papandreou, F.~Schroff, {\em et~al.}, ``Rethinking atrous convolution for semantic image segmentation,'' {\em ArXiv} {\bf abs/1706.05587}  (2017).

\bibitem{dutta2019assessment}
S.~Dutta, B.~Das, and S.~Kaushik, ``{Assessment of optimal deep learning configuration for vertebrae segmentation from {CT} images},'' in {\em Medical Imaging 2019: Imaging Informatics for Healthcare, Research, and Applications},   {\bf 10954}, 298 -- 305, SPIE  (2019).

\bibitem{lin2017focal}
T.-Y. Lin, P.~Goyal, R.~Girshick, {\em et~al.}, ``Focal loss for dense object detection,'' in {\em Proceedings of the IEEE International Conference on Computer Vision (ICCV)},   (2017).

\bibitem{ronneberger2015unet}
O.~Ronneberger, P.~Fischer, and T.~Brox, ``U-net: Convolutional networks for biomedical image segmentation,'' in {\em Medical Image Computing and Computer-Assisted Intervention -- MICCAI 2015},  N.~Navab, J.~Hornegger, W.~M. Wells, {\em et~al.}, Eds., 234--241, Springer International Publishing, (Cham)  (2015).

\bibitem{protection2007icrp}
ICRP, ``The 2007 recommendations of the international commission on radiological protection: {ICRP} publication 103,'' {\em Ann ICRP} {\bf 37}, 1--332  (2007).

\bibitem{mccollough2014use}
C.~McCollough, D.~M. Bakalyar, M.~Bostani, {\em et~al.}, ``Use of water equivalent diameter for calculating patient size and size-specific dose estimates (ssde) in {CT}: the report of aapm task group 220,'' {\em AAPM report} {\bf 2014}, 6  (2014).

\bibitem{wang2021region}
Y.~Wang, B.~Liu, F.~Zhou, {\em et~al.}, ``Region context aggregation network for multi-organ segmentation on abdominal ct,'' in {\em International Conference on Image and Graphics},  664--674, Springer  (2021).

\bibitem{lin2021variance}
H.~Lin, Z.~Li, Z.~Yang, {\em et~al.}, ``Variance-aware attention u-net for multi-organ segmentation,'' {\em Medical Physics}   (2021).

\bibitem{aapm2019size}
AAPM, ``Size-specific dose estimate (ssde) for head {CT},'' {\em American Association of Physicists in Medicine (AAPM) Report No. 293}   (2019).

\bibitem{juszczyk2021automated}
J.~Juszczyk, P.~Badura, J.~Czajkowska, {\em et~al.}, ``Automated size-specific dose estimates using deep learning image processing,'' {\em Medical Image Analysis} {\bf 68}, 101898  (2021).

\bibitem{schmidt2016accuracy}
T.~G. Schmidt, A.~S. Wang, T.~Coradi, {\em et~al.}, ``Accuracy of patient-specific organ dose estimates obtained using an automated image segmentation algorithm,'' {\em Journal of Medical Imaging} {\bf 3}(4), 043502  (2016).

\bibitem{kalender2014generating}
W.~A. Kalender, N.~Saltybaeva, D.~Kolditz, {\em et~al.}, ``Generating and using patient-specific whole-body models for organ dose estimates in ct with increased accuracy: feasibility and validation,'' {\em Physica Medica} {\bf 30}(8), 925--933  (2014).

\bibitem{tian2016convolution}
X.~Tian, W.~P. Segars, R.~L. Dixon, {\em et~al.}, ``Convolution-based estimation of organ dose in tube current modulated ct,'' {\em Physics in Medicine \& Biology} {\bf 61}(10), 3935  (2016).

\bibitem{imran2022multimodal}
A.-A.-Z. Imran, S.~Wang, D.~Pal, {\em et~al.}, ``Multimodal contrastive learning for prospective personalized estimation of ct organ dose,'' in {\em International Conference on Medical Image Computing and Computer-Assisted Intervention},  634--643, Springer  (2022).

\end{thebibliography}
\bibliographystyle{spiejour}   

\end{spacing}
\end{document}